\documentclass[10pt]{article}
\PassOptionsToPackage{type1}{cabin}
\usepackage[citestyle=numbers]{colab}
\usepackage{flafter}
\usepackage{placeins}
\usepackage{needspace}
\usepackage{xurl}

\newcommand{\uavtablestyle}{%
  \small
  \setlength{\tabcolsep}{3pt}%
  \renewcommand{\arraystretch}{1.1}%
}
\newcolumntype{Y}{>{\raggedright\arraybackslash}X}
\newcommand{\supptablestyle}{%
  \small\renewcommand{\arraystretch}{1.13}\setlength{\tabcolsep}{6pt}%
}

\makeatletter
\apptocmd{\thebibliography}{\clubpenalty=10000\@clubpenalty=10000\widowpenalty=10000}{}{}
\makeatother
\newcommand{\uavpapertitle}{UAV-Flow++: A Unified Suite for Benchmarking and Learning Language-Guided UAV Flight}
\title{\uavpapertitle}
\author{%
  Donglin Yang$^{1}$ \quad Xiangyu Wang$^{1}$ \quad Yantong Zhong$^{1}$ \quad Wenhao Zheng$^{1}$ \\[3pt]
  Zihan Ding$^{1}$ \quad Yue Liao$^{2}$ \quad Hongsheng Li$^{3}$ \quad Ke Xu$^{1}$ \quad Si Liu$^{1,*}$ \\[6pt]
  {\small $^{1}$Beihang University, Beijing, China 100191} \\[2pt]
  {\small $^{2}$National University of Singapore, Singapore 119077} \\[2pt]
  {\small $^{3}$MMLab, The Chinese University of Hong Kong, Shatin, New Territories, Hong Kong, China} \\[3pt]
  {\small $^{*}$Corresponding author: Si Liu (\texttt{liusi@buaa.edu.cn})}
}
\labname{Colab}
\institution{}
\colabdate{}
\paperurl{https://prince687028.github.io/UAV-Flow}
\githuburl{https://github.com/buaa-colalab/UAV-Flow}
\huggingfaceurl{}
\dataurl{https://huggingface.co/datasets/wangxiangyu0814/UAV-Flow}
\colabrunningtitle{\uavpapertitle}
\hypersetup{
  pdftitle={\uavpapertitle},
  pdfauthor={Donglin Yang, Xiangyu Wang, Yantong Zhong, Wenhao Zheng, Zihan Ding, Yue Liao, Hongsheng Li, Ke Xu, Si Liu}
}

\begin{document}
\maketitle
\vspace{-0.3cm}

\begin{colababstract}
Language-guided UAV navigation has largely focused on long-horizon route planning, leaving short-range, object-relative flight maneuvers underexplored. To address this gap, we formulate Flying-on-a-Word (Flow), a task requiring a UAV to convert concise natural-language instruction, egocentric visual observation, and current state dynamics into continuous flight actions. Flow encompasses both primitive motion commands and complex object-interactive behaviors, assessing motion intent comprehension and spatial context grounding. To enable policy learning and closed-loop evaluation, we introduce UAV-Flow, a comprehensive dataset comprising 30,692 expert-piloted real-world trajectories across eight distinct motion categories, alongside 10,109 simulation trajectories in the UAV-Flow-Sim environment equipped with task-specific completion criteria. Furthermore, we develop FlowVLA, which undergoes imitation pretraining on mixed real-sim demonstrations and online post-training via Tail-Aware GRPO (TGRPO). By leveraging group reward distributions, TGRPO effectively reinforces high-reward action sequences. Evaluated on 273 held-out UAV-Flow-Sim episodes, pretrained FlowVLA achieves a mean success rate of 75.58\%, outperforming the strongest baseline by 6.72 percentage points, which TGRPO post-training further elevates to 93.16\% with superior sample efficiency. Physical UAV experiments demonstrate robust execution of object-relative commands in both indoor and outdoor real-world scenes. Together, Flow establishes a novel benchmark and learning framework for language-conditioned flight maneuvers.

\medskip
\colablinks
\end{colababstract}

\begin{figure}[!htbp]
\centering
\includegraphics[width=0.98\textwidth]{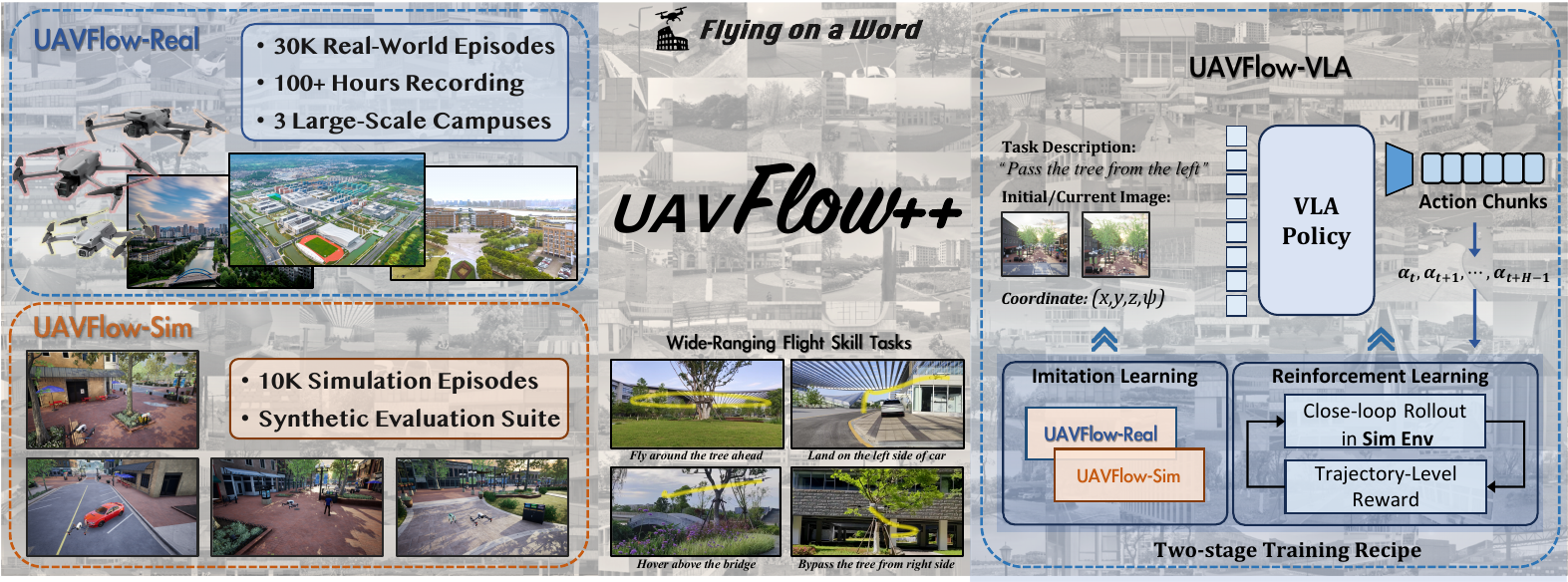}
\caption{\textbf{Overview of UAV-Flow++.} UAV-Flow++ comprises real-world flight demonstrations, an interactive simulation suite, and a vision-language-action policy for language-conditioned UAV maneuvers. The real-world dataset pairs language instructions with onboard observations, UAV states, and expert actions across primitive motion and object-interactive tasks. UAV-Flow-Sim supports closed-loop policy training and evaluation. FlowVLA predicts position-yaw action chunks from language instructions, initial and current images, and UAV states, using imitation pretraining on mixed real and simulated demonstrations followed by reinforcement learning with trajectory-level rewards.}
\label{fig:uavflow_overview}
\end{figure}

\section{Introduction}

Unmanned Aerial Vehicles (UAVs) are widely deployed across diverse applications, serving as aerial perception platforms for object detection, tracking, counting, and human behavior understanding~\cite{visdrone,uav123,animaldrone,droneswarm,uavhuman}, as well as enabling precision agriculture, infrastructure inspection, emergency response, and entertainment applications~\cite{blei2025cloudtrack, au2025contingency, wang2026Skyfind, yao2024aeroverse}.
In parallel, advances in motion planning and control have enabled increasingly autonomous flight and multi-UAV coordination in cluttered three-dimensional environments~\cite{uavmotion,ego_planner,ego_swarm,usenko2017realtime}. These capabilities provide a technical foundation for UAVs to operate as embodied agents that interpret human intent and execute tasks in the physical world.
While modern UAVs are increasingly capable of autonomous flight, enabling them to follow high-level natural language instructions remains a fundamental challenge in achieving more intelligent human--UAV interaction. For example, executing a command such as \emph{``Fly 3 meters in front of me and take a photo"} requires UAVs to transform abstract human intentions into executable physical actions. 
This process demands that the system seamlessly interpret user intent, reason about the environment, and synthesize dynamically feasible, collision-free flight trajectories.

In pursuit of language-guided aerial autonomy, recent efforts have extended vision-language navigation (VLN) paradigms from ground robots to UAV platforms \cite{anderson2018r2r,Chen_2022_duet,cma,Liu_2023_Aerialvln,lee2024citynav,gao2025openfly,wang2024openuav}. Leveraging large-scale simulated environments \cite{airsim,unrealcv,gazebo,zhong2024unrealzoo,yao2024aeroverse}, these approaches enable UAVs to follow language instructions for long-horizon target search and goal-directed navigation tasks. 
However, prevailing research mainly prioritizes high-level semantic reasoning and macro-level route synthesis
\cite{anderson2018r2r,Chen_2022_duet,ku2020rxr,zhou2024navgpt,zhou2024navgpt2,lin2025navcot}
, operating on the premise that short-range execution can be offloaded to standard low-level planners and control modules \cite{uavmotion,ego_planner,ego_swarm}. While this assumption has proven successful in many robotic systems, extending it to UAV platforms introduces additional challenges. UAVs must plan and execute dynamically feasible motions while avoiding obstacles in unconstrained 3D environments under real-world kinematic constraints \cite{uavmotion,ego_planner,usenko2017realtime}. As a result, enabling language-guided low-level flight maneuver remains a critical yet largely underexplored problem in the development of intelligent UAV systems.

We introduce the \emph{Flying-on-a-Word (Flow)} task to formalize language-conditioned short-range maneuver execution. Given a concise instruction, egocentric observations, and UAV state estimates, the agent must interpret the intended motion and execute the corresponding flight behavior through closed-loop interaction. Flow covers primitive motion commands and object-interactive commands, assessing motion intent understanding and spatial context grounding. 
To support systematic evaluation and policy learning on language-conditioned short-range UAV flight, we present \emph{UAV-Flow}, the first large-scale benchmark that integrates real-world expert demonstrations with an interactive simulation suite. Specifically, \emph{UAV-Flow} comprises 30,692 expert-piloted real-world trajectories spanning eight major motion categories and diverse visual contexts. Complementing these demonstrations, \emph{UAV-Flow-Sim} provides 10,109 simulated trajectories in an interactive environment, enabling controlled policy training and closed-loop evaluation. 
We further develop an automated evaluation method for UAV-Flow-Sim to assess task success by jointly considering maneuver execution quality and final state instruction compliance.

Next, we introduce a staged policy learning framework that jointly exploits expert demonstrations and simulation-based interaction to make effective use of task-specific data. In the first stage, imitation learning is used to ground the policy in the distribution of expert flight behaviors, enabling the model to acquire instruction-compliant maneuvers from offline demonstrations. However, during closed-loop execution, control deviations can lead the policy to states outside the demonstration distribution. 
We further improve the policy through environment interaction in UAV-Flow-Sim, using task-level completion feedback to guide policy optimization.
Guided by this task-level feedback, we seek to learn more effectively from the action sequences of trajectories with higher completion scores. Therefore, we build upon GRPO~\cite{shao2024deepseekmath} and introduce a tail-aware advantage derived from the group reward distribution, producing tail-aware GRPO (TGRPO). This advantage construction assigns greater weight to reward intervals reached by fewer trajectories within each group, strengthening learning from high-reward execution while making fuller use of outcomes that are sparsely represented in group.
For physical deployment, we use a ground-drone collaborative architecture in which the ground station performs VLA inference and the UAV executes the commands. The execution layer selects look-ahead waypoints according to the measured end-to-end delay, accounting for UAV motion during communication and inference to support continuous closed-loop flight.

Our contributions are summarized as follows:

(1) We define the \emph{Flying-on-a-Word (Flow)} task, which formulates language-conditioned UAV navigation as fine-grained, short-range trajectory execution from egocentric observations, assessing the policy's ability to understand motion intent and execute object-grounded spatial behaviors.

(2) We build \emph{UAV-Flow}, a real-world expert demonstration dataset, and \emph{UAV-Flow-Sim}, an interactive simulation suite, together forming the large-scale real-world benchmark for language-conditioned UAV short-range navigation. We also provide an automated task-level closed-loop evaluation protocol in the simulation benchmark.

(3) We propose a staged policy learning framework that combines imitation learning and reinforcement learning through imitation pretraining on mixed real-world and simulated demonstrations, followed by closed-loop post-training with Tail-Aware GRPO (TGRPO).

(4) We have conducted extensive experiments. The pretrained FlowVLA policy achieves a success rate of 75.58\%, which TGRPO post-training further raises to 93.16\%, achieving the highest overall success rate among evaluated methods. Comprehensive ablation studies validate the effectiveness of the policy learning framework.

This paper is built upon \emph{UAVFlow}~\cite{wang2026uav}, our conference version, and extends it in several important aspects. First, we further develop an upgraded closed-loop simulation evaluation suite, \emph{UAV-Flow-Sim}, equipped with newly introduced automated task-level evaluation metrics for more fine-grained and reproducible assessment. Second, we develop FlowVLA with mixed-demonstration pretraining followed by online RL post-training, incorporating a tail-aware advantage into GRPO to strengthen learning from the action sequences of high-reward trajectories and improve success rates on challenging \emph{Flow} tasks. Third, we provide a much more comprehensive experimental study, including comparisons with prior methods, ablation studies, and qualitative analyses. Compared with the conference version, the extended method achieves clear performance improvements and provides a more complete benchmark, training, and evaluation framework for language-conditioned UAV maneuvers.

\section{Related Work}

\subsection{Datasets and Benchmarks}

Language-guided UAV research is evolving from vision-and-language navigation (VLN) toward vision-language-action (VLA) learning, while gradually extending from simulation-based studies toward real-world data. Early benchmarks primarily focus on long-horizon route or goal navigation in simulated environments. AVDN~\cite{fan2023aerial} investigates dialog-guided flight, while CityNav~\cite{lee2024citynav} extends goal-oriented navigation to reconstructed urban environments. AerialVLN~\cite{Liu_2023_Aerialvln} further studies long-horizon language-guided flight in city-scale simulations using a discrete 4-DoF action space. 
Subsequent works explore more diverse task and action settings: OpenUAV~\cite{wang2024openuav} introduces continuous trajectories, OpenFly~\cite{gao2025openfly} scales up simulator-based aerial VLN, IndoorUAV~\cite{liu2025indooruav} combines long-horizon VLN with short-horizon VLA data, and HUGE-Bench~\cite{guo2026hugebench} evaluates high-level multi-stage VLA tasks. As shown in Table~\ref{tab:uav_benchmarks}, these benchmarks cover increasingly diverse tasks and action spaces, but the dataset and environment interactions remain primarily simulation-based.
Beyond navigation, AeroVerse~\cite{yao2024aeroverse} broadens the scope of aerial embodied intelligence to include scene awareness, spatial reasoning, navigational exploration, task planning, and motion decision-making. 
Meanwhile, existing real-world UAV datasets mainly focus on visual perception~\cite{visdrone,uav123,animaldrone,uavhuman}, with relatively few providing demonstrations that align language instructions, visual observations, UAV states, and executable actions for policy learning.
UAV-Flow~\cite{wang2026uav} addresses this gap by providing real-world language-vision-action demonstrations for fine-grained flight control learning, together with UAV-Flow-Sim for reproducible closed-loop evaluation. This combination advances language-guided UAV research beyond simulation-based navigation toward learning and evaluating action policies grounded in real-world flight demonstrations.

\begin{table}[!htbp]
\centering
\uavtablestyle
\caption{\textbf{Comparison of language-guided UAV benchmarks.}}
\label{tab:uav_benchmarks}
\begin{tabular*}{\textwidth}{@{\extracolsep{\fill}}lccccc@{}}
\toprule
\textbf{Benchmark}
& \textbf{Action Type}
& \textbf{\#DoF}
& \textbf{\#Traj}
& \textbf{Collection Setting}
& \textbf{Task Granularity} \\
\midrule
AVDN~\cite{fan2023aerial}
& Continuous
& 3
& 3K
& Sim
& Long-horizon dialog navigation \\
AerialVLN~\cite{Liu_2023_Aerialvln}
& Discrete
& 4
& 8K
& Sim
& Long-horizon navigation \\
CityNav~\cite{lee2024citynav}
& Discrete
& 4
& 32K
& Sim
& Goal-directed navigation \\
OpenUAV~\cite{wang2024openuav}
& Continuous
& 6
& 12K
& Sim
& Goal-directed navigation \\
OpenFly~\cite{gao2025openfly}
& Discrete
& 4
& 100K
& Sim
& Long-horizon navigation \\
IndoorUAV~\cite{liu2025indooruav}
& Discrete
& 4
& 50K
& Sim
& Long-horizon VLN / short-horizon VLA \\
HUGE-Bench~\cite{guo2026hugebench}
& Continuous
& 4
& 6K
& Sim
& High-level multi-stage VLA \\
\midrule
\textbf{UAV-Flow}
& Continuous
& 6
& 40K
& Sim + Real
& Short-range execution \\
\bottomrule
\end{tabular*}
\end{table}

\subsection{End-to-End Vision-Language-Action Models}

End-to-end VLA models map visual observations and language instructions to executable actions. OpenVLA~\cite{kim2024openvla} autoregressively predicts discretized action tokens, whereas $\pi_0$~\cite{black2024pi0} and $\pi_{0.5}$~\cite{physicalintelligence2025pi05} employ flow-matching action experts conditioned on pretrained VLM features to generate continuous action chunks. Building on these architectures, OpenVLA-OFT~\cite{kim2025openvlaoft} improves action generation and control efficiency through action chunking, parallel decoding, and continuous action regression. FAST~\cite{pertsch2025fast} further improves the efficiency of autoregressive action prediction by compressing action chunks using discrete cosine transforms and byte-pair encoding. These methods leverage pretrained VLMs to jointly encode visual observations and language instructions, providing representations for language-conditioned action prediction.
Recent work further improves pretrained VLAs through reinforcement learning, through feedback collected during policy execution. VLA-RL~\cite{lu2025vlarl} introduces trajectory-level online optimization, while SimpleVLA-RL~\cite{li2026simplevlarl} supports scalable training through VLA-specific rollout sampling and parallel execution. RIPT-VLA~\cite{tan2025riptvla} uses sparse success rewards with dynamic sampling and leave-one-out advantage estimation. EXPO-FT~\cite{dong2026expoft} combines an action-editing policy with Q-guided candidate selection for sample-efficient adaptation in real-world manipulation.

VLA models are also being extended to aerial robots. IndoorUAV combines indoor VLN with short-horizon $\pi_0$-based execution~\cite{liu2025indooruav}, while CognitiveDrone~\cite{lykov2025cognitivedrone} and RaceVLA~\cite{serpiva2025racevla} use velocity commands in simulation and on physical UAVs, respectively. AutoFly~\cite{sun2026autofly} integrates language guidance with obstacle-aware navigation. AirVLA~\cite{tucker2026airvla} transfers manipulation-pretrained VLAs to aerial navigation and pick-and-place tasks, combining demonstrations synthesized with gaussian splatting and payload-aware inference-time guidance to accommodate flight dynamics. For language-conditioned UAV navigation, we adopt a two-stage VLA training framework: imitation pretraining on mixed real-world and simulated demonstrations, followed by online reinforcement learning through interaction in simulation.

\FloatBarrier
\section{UAV-Flow Benchmark}
\label{sec:benchmark}

This section introduces UAV-Flow, a benchmark for evaluating language-conditioned short-range UAV flight execution. The benchmark is centered on the Flying-on-a-Word (Flow) task, in which a UAV interpret a concise natural-language instruction and execute the corresponding fine-grained flight behavior based on onboard visual observations and proprioceptive feedback. In contrast to existing aerial vision-language navigation benchmarks, which primarily focus on high-level navigation and long-horizon decision making, UAV-Flow targets the execution of short-range, instruction-grounded maneuvers. It evaluates whether an agent can infer intended motion, ground spatial references in the current scene, and generate physically feasible UAV behaviors. These capabilities are essential for precise UAV operation yet are largely underexplored in existing benchmarks. To facilitate both policy learning and reproducible benchmarking, UAV-Flow provides expert-piloted real-world demonstrations for training with a complementary simulation suite that enables large-scale closed-loop evaluation under controlled conditions.

\subsection{Task Formulation}
\label{sec:task_formulation}

\begin{figure}[!t]
\centering
\includegraphics[width=0.98\textwidth]{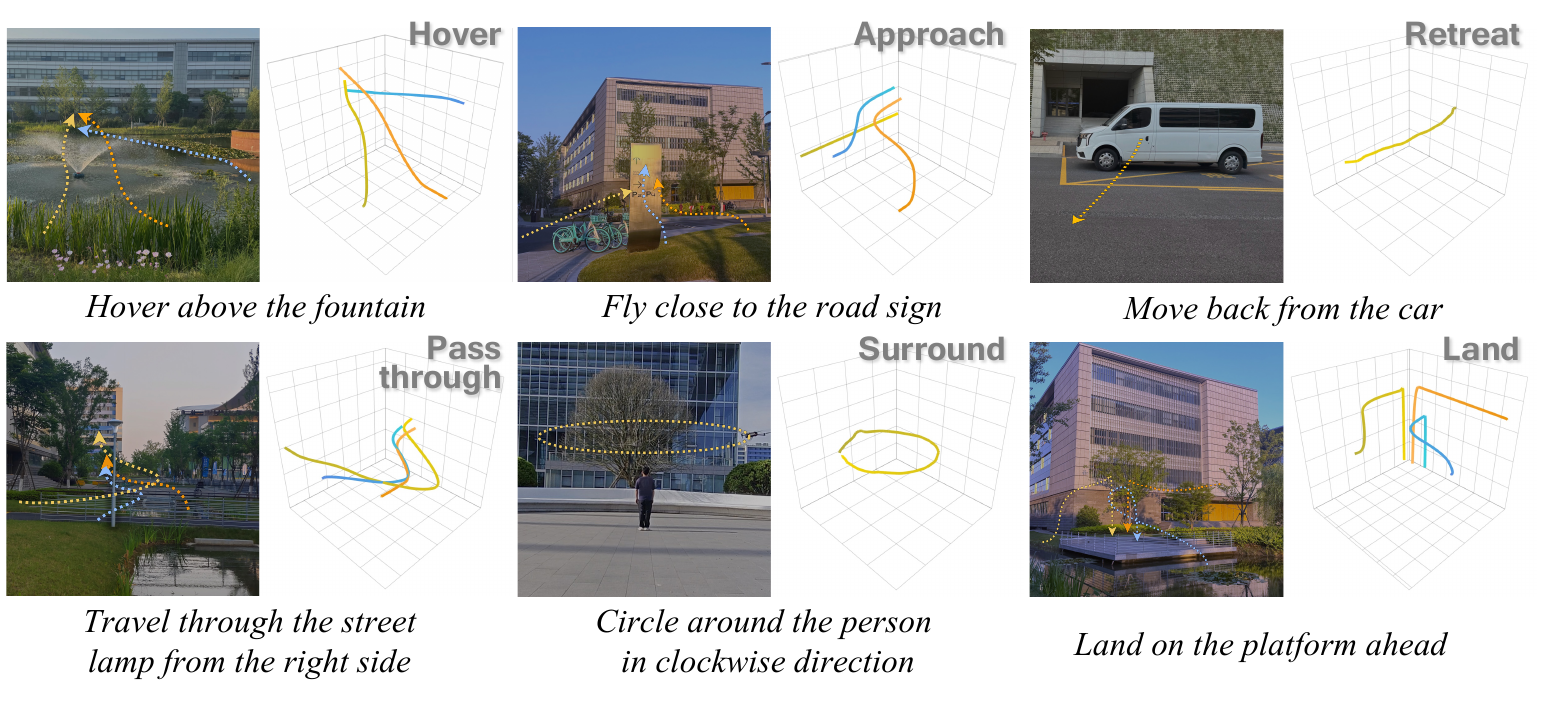}
    \caption{\textbf{Representative Flow tasks.} Examples show scene observations, natural-language instructions, and corresponding 3D flight trajectories for hovering, approaching, retreating, passing, orbiting, and landing. Multiple expert trajectories for the same instruction illustrate the diversity of demonstrated flight behaviors.}
    \label{fig:flow_task_examples}
\end{figure}

To study language-conditioned fine-grained UAV control, we introduce the \emph{Flying-on-a-Word} (Flow) task setting, which aligns concise language instructions with short-range flight execution. Flow emphasizes visually grounded interactions and simple expert-pilot maneuvers, such as orbiting or passing around obstacles, rather than long-horizon route planning. Each Flow instance provides the UAV agent with three modalities at every time step: a natural-language instruction $\mathbf{i}$, the UAV 6-DoF state $\mathbf{s}_t$ expressed in an episode-local trajectory frame, and an egocentric visual observation $\mathbf{o}_t$. The policy is expected to generate continuous UAV actions that reflect the instruction intent while satisfying dynamic feasibility:
\begin{equation}
\pi_\theta: (\mathbf{o}_t,\mathbf{s}_t,\mathbf{i}) \mapsto \mathbf{a}_t .
\label{eq:flow_policy}
\end{equation}
The action $\mathbf{a}_t$ is represented as a target pose or a position-yaw command in a local coordinate frame, which can be mapped to platform-specific flight controllers. Executing the policy over an episode of length $T$ produces an action sequence $A={\mathbf{a}_1,\mathbf{a}_2,\ldots,\mathbf{a}_T}$ and a state trajectory $\tau={\mathbf{s}_1,\mathbf{s}_2,\ldots,\mathbf{s}_T}$. A successful Flow trajectory should faithfully capture and execute the motion semantics conveyed by the instruction, align with spatial references in the observed scene, lead to the intended final state, and comply with UAV flight constraints.

The Flow task is designed to evaluate two fundamental capabilities required for language-conditioned UAV flight execution. The first is \emph{motion intent understanding}, which requires the policy to infer the desired flight behavior specified by the instruction, such as translating in a particular direction, rotating by a given angle, ascending, hovering, or landing. 
The second is \emph{spatial context grounding}, which requires the policy to map spatial references in natural-language instructions to corresponding objects or locations in visual observations and generate actions accordingly. Examples include approaching a vehicle, passing an object on a specified side, or stopping near a landmark. 
Together, these capabilities enable the agent to map language instructions into physically feasible flight actions that faithfully capture the intended motion semantics and respect the spatial relationships in the observed scene.

To evaluate these capabilities systematically, UAV-Flow organizes instructions into two categories: primitive motion commands and object-interactive commands. The former primarily assess the agent's ability to understand and execute motion intent, while the latter further require grounding language in the surrounding scene and performing actions relative to referenced entities. The real-world dataset organizes expert demonstrations into eight major motion categories for statistical analysis, whereas UAV-Flow-Sim instantiates the same task formulation in simulation, replicating the real-world setup. This design disentangles instruction-level motion semantics from the scene entities and spatial relations that govern execution. Fig.~\ref{fig:flow_task_examples} illustrates representative Flow trajectories.

\subsection{Real-World Dataset}
\label{sec:real_world_dataset}

\begin{figure}[!htbp]
\centering
    \includegraphics[width=0.98\textwidth]{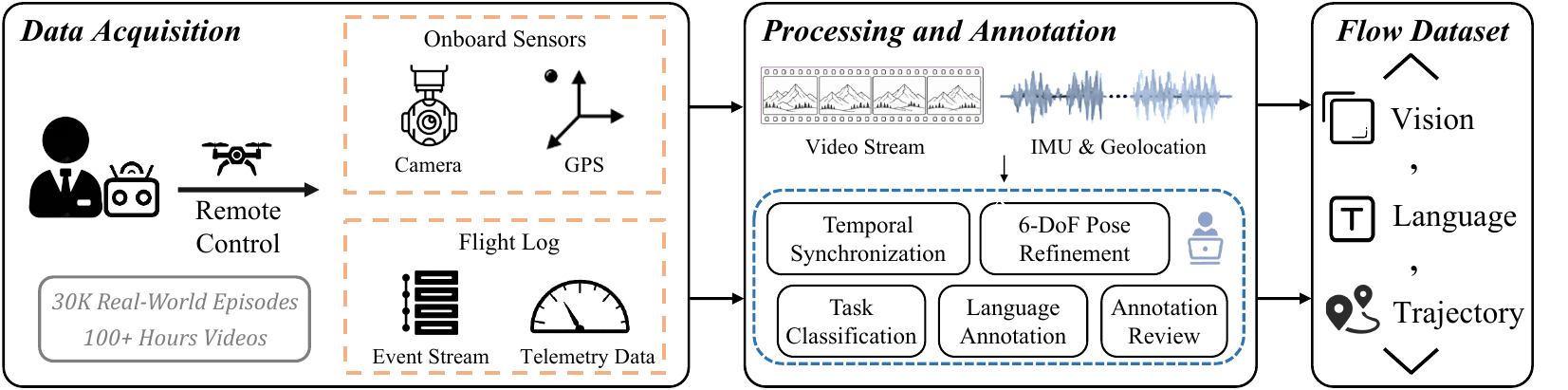}
    \caption{\textbf{Overview of real-world UAV-Flow dataset construction.}
    Expert pilots collect onboard videos and flight logs, followed by temporal synchronization and 6-DoF pose registration. Each flight demonstration is assigned a task label and a language instruction, followed by annotation review. The resulting UAV-Flow dataset pairs egocentric videos and flight trajectories with the corresponding instructions.}
    \label{fig:uavflow_pipeline}
    \label{fig:datasetpipeline}
\end{figure}

To capture expert flight behaviors, we construct UAV-Flow by collecting and annotating real-world flight demonstrations. Fig.~\ref{fig:datasetpipeline} summarizes the data collection and annotation pipeline.

\textbf{Trajectory acquisition.}
Data acquisition is conducted across three university campuses covering 5.02\,km\textsuperscript{2}. These campuses provide compact but diverse outdoor environments with pedestrians, vehicles, vegetation, buildings, roads, open spaces, and landmarks. Flights are manually operated by certified UAV pilots, each with over 800 hours of flight experience, using commercial-grade platforms such as the DJI Mavic 3T RTK equipped with 4K cameras and RTK GPS modules. These sensors provide high-resolution first-person-view videos and centimeter-level trajectory accuracy for trajectory reconstruction and multimodal alignment. To guide flight execution, we curate instruction templates spanning primitive motion commands (\emph{e.g.}, takeoff, shift, and rotate) and object-interactive behaviors (\emph{e.g.}, orbiting around a car or hovering beside a landmark). Pilots execute these instructions exclusively from the UAV's onboard view, so the recorded visual stream matches the visual input available to a deployed policy. To increase behavioral coverage, each instruction is executed from multiple initial positions and viewing directions. Thus, semantically equivalent commands can produce diverse trajectories while preserving the same underlying intent.

\textbf{6-DoF Pose Registration.}
Each demonstration contains synchronized onboard video and complete flight logs. Using precise timestamps, we temporally synchronize the two data streams by linearly interpolating each 6-DoF UAV state, including GPS coordinates and local orientation, at each timestamp of the associated video frame. This establishes a frame-wise correspondence between each visual observation and its concurrent UAV state. We then register the synchronized poses to an episode-local coordinate frame: GPS coordinates are transformed into a local cartesian frame centered at the trajectory start, while orientations are represented relative to the initial heading. Expert actions are derived from the synchronized flight logs and registered pose sequences, yielding pose commands consistent with the action used for policy learning. The aligned sequences are sampled at 5 Hz, producing temporally consistent tuples of visual observation, UAV state, and expert action.

\textbf{Instruction annotation.}
Language annotations are constructed through a quality-controlled process. Annotators first filter clips with ambiguous behaviors, unsafe execution, unclear scene relations, or inconsistent motion semantics. They then assign concise commands that describe the intended motion and, when needed, the spatial relation to a visible object. To support standardized annotation and efficient data processing, UAV-Flow adopts a fixed command set for canonical task labels, while constructing an open-vocabulary command set by paraphrasing the canonical annotations with large language models~\cite{achiam2023gpt}. The resulting real-world dataset contains 30,692 trajectories across 8 major motion categories, paired with egocentric videos, aligned UAV states, and instructions.

\subsection{Simulation Suite}
\label{sec:simulation_suite}
UAV-Flow-Sim complements the real-world dataset by providing additional trajectories and a controlled environment for closed-loop policy evaluation. Within this environment, policies can be compared under identical initial states, scene layouts, and termination criteria. 

The simulator is built in a campus scene and uses UnrealCV~\cite{unrealcv,zhong2024unrealzoo} for rendering and UAV control. The UAV is controlled through a position-yaw interface that approximates the position-mode control used by practical UAV systems. The environment includes static landmarks and interactive objects such as humans, vehicles, and quadruped robots, enabling both primitive motion tasks and object-interactive tasks. 
We collect the UAV-Flow-Sim dataset through a hybrid strategy that combines human piloting with rule-based trajectory generation. Human pilots manually execute complex flight instructions, while rule-based scripts handle instructions specifying fixed flight maneuvers. These scripts use privileged simulator information, including UAV states and object locations, to plan and execute \mbox{trajectories} for the prescribed maneuvers. 
The dataset follows the same instruction, visual observation, and state format as the real-world data and contains 10,109 trajectories. 

\textbf{Closed-loop evaluation protocol.}
Each episode is initialized with a designated UAV pose and task instruction. At each step, the simulator feeds the current egocentric observation and UAV state to the policy, which in turn predicts an action. The simulator then executes this action and updates the environment before returning the subsequent observation. An episode terminates upon task completion, collision, safety violation, or horizon expiration. 
The suite records the complete executed trajectory, terminal state, object relations, and collision events. These records allow us to examine how prediction errors affect subsequent flight behavior and whether policies recover from the resulting deviations during execution.

\textbf{Task completion score.}
We upgrade the closed-loop evaluation protocol of UAV-Flow-Sim by introducing an automated task completion score. For each task type $k$, an episode receives a normalized completion score
\begin{equation}
C_k(\tau) = C_{\mathrm{proc}}^k(\tau) + C_{\mathrm{goal}}^k(\tau),
\quad C_k(\tau) \in [0,1],
\label{eq:completion_score}
\end{equation}
where $C_{\mathrm{proc}}^k$ measures execution of the intended maneuver and $C_{\mathrm{goal}}^k$ measures whether the final UAV state satisfies the target condition. Both terms use task-specific predicates or tolerance scores. The completion score provides a graded measure of task fulfillment, reflecting both maneuver execution and terminal-state accuracy.

For example, in a passing task, $C_{\mathrm{proc}}^k$ evaluates whether the UAV traverses the required passage or passes the specified object or region during execution, while $C_{\mathrm{goal}}^k$ checks whether it stops at the target position. For rotation tasks, the score evaluates whether the UAV follows the commanded direction and achieves the required yaw change, accounting for deviations in terminal position and orientation. Object-interactive tasks additionally assess object distance, relative side, facing direction, or landing proximity, as required by the instruction.

For distance- or angle-based terms, we use a clipped interval score
\begin{equation}
\Phi(d;\Delta,w)
= \max\left(0, w - 0.1 \left\lceil \frac{d}{\Delta} \right\rceil \right),
\label{eq:completion_phi}
\end{equation}
where $d$ is the corresponding error, $\Delta$ is the error interval associated with each 0.1-point score reduction, and $w$ is the maximum weight of the term. 

\begin{table}[!t]
\centering
\uavtablestyle
\caption{\textbf{Trajectory distribution across action types in UAV-Flow and UAV-Flow-Sim.} Percentages indicate the proportion of trajectories within each dataset.}
\label{tab:uavflow_statistics}
\begin{tabular*}{0.6\columnwidth}{@{\extracolsep{\fill}}cccc@{}}
\toprule
\textbf{Category} & \textbf{Action Type} & \textbf{Real} & \textbf{Sim} \\
\midrule
\multirow{4}{*}{\shortstack{Primitive\\Motion}}   & Shift       & 4.3k (13.9\%) & 3.0k (29.4\%) \\
                         & Rotate      & 1.8k (5.9\%)  & 1.0k (10.3\%) \\
                         & Hover       & 3.2k (10.6\%) & 0 (0\%)      \\
                         & Retreat     & 1.7k (5.5\%)  & 1.3k (12.6\%) \\
\midrule
\multirow{4}{*}{\shortstack{Object\\Interaction}}     & Surround    & 1.0k (3.4\%)  & 0.2k (2.3\%)  \\
                         & Approach    & 8.6k (28.1\%) & 2.6k (25.6\%) \\
                         & Pass through& 8.7k (28.4\%) & 1.5k (15.2\%) \\
                         & Land        & 1.3k (4.3\%)  & 0.5k (4.7\%)  \\
\midrule
\textbf{Total} & \textendash{} & \textbf{30.7k} & \textbf{10.1k} \\
\bottomrule
\end{tabular*}
\end{table}

\textbf{Trajectory similarity.}
Completion score measures whether a task is solved, while trajectory similarity characterizes agreement with a reference flight path. 
We further assess trajectory similarity using an adapted normalized dynamic time warping (NDTW) score~\cite{ndtw}, which compares the executed trajectory $\tau$ with the episode reference trajectory $\tau^*$:
\begin{equation}
\mathrm{NDTW}(\tau,\tau^*) =
\exp\left(-\frac{\mathrm{DTW}(\tau,\tau^*)}{L(\tau^*)}\right),
\label{eq:ndtw}
\end{equation}
where both trajectories are expressed in the same episode-local coordinate frame and $L(\tau^*)$ is the reference path length in the representation defined below. We represent each trajectory point as a six-dimensional vector by concatenating its position with the cosine values of its roll, yaw, and pitch angles:
\begin{equation}
\mathbf{x}_t = [x_t,y_t,z_t,\cos\phi_t,\cos\psi_t,\cos\theta_t],
\label{eq:trajectory_state}
\end{equation}
where $x_t,y_t,z_t$ denote position in meters, and $\phi_t,\psi_t,\theta_t$ denote roll, yaw, and pitch, respectively.

\Needspace{6\baselineskip}
\subsection{Data Analysis}
\label{sec:data_analysis}

UAV-Flow and UAV-Flow-Sim provide complementary coverage for Flow formulation. The real-world dataset contains 30,692 expert-piloted trajectories across 8 major motion categories, capturing diverse visual contexts, human control styles, and physically executed UAV dynamics. The simulation dataset contains 10,109 trajectories generated under matched instruction and state formats, enabling controlled testing and simulation-based policy optimization. We further construct a simulation test set comprising 273 annotated trajectories, covering all major action types in the dataset to facilitate systematic evaluation on the Flow task. Each test episode specifies its task label, initial state, instruction, reference trajectory, and task-specific completion criteria, establishing a reproducible protocol for closed-loop evaluation.

Table~\ref{tab:uavflow_statistics} summarizes the trajectory distribution across action types in UAV-Flow and UAV-Flow-Sim. Approaching and passing account for the largest shares of the real-world dataset, whereas shifting and approaching are the most frequent actions in simulation. The distributions differ across the two datasets; in particular, hovering trajectories are present in the real-world data but absent from the simulation data. 
Compared with existing aerial VLN datasets such as AerialVLN~\cite{Liu_2023_Aerialvln}, which prioritize long-range paths, discrete actions, and high-level route planning, UAV-Flow provides continuous expert trajectories and a closed-loop simulation suite tailored for short-range, language-conditioned motion execution and fine-grained spatial interaction. Reflecting this emphasis on precision and agility, most trajectories in UAV-Flow span under 20 meters in length.

\section{Method}
\label{sec:method}

\begingroup
\setlength{\columnsep}{0.03\textwidth}
\setlength{\intextsep}{0pt}
\begin{wrapfigure}{r}{0.5\textwidth}
\centering
\includegraphics[width=0.93\linewidth]{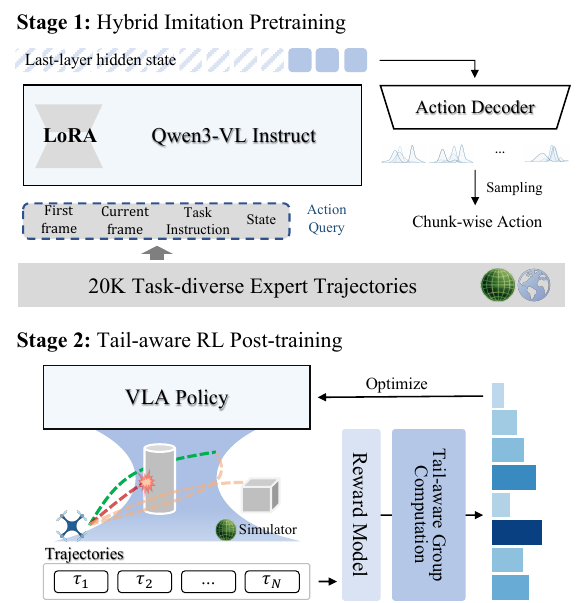}
\caption{\textbf{Overview of the two-stage training pipeline.}
Stage one trains a language-conditioned UAV policy through imitation
learning on mixed real-world and simulated demonstrations. In stage two, the policy undergoes interactive reinforcement learning via TGRPO, leveraging simulated closed-loop rollouts and tail-aware advantages derived from task completion scores for policy optimization.}
\label{fig:method}
\end{wrapfigure}
This section presents FlowVLA's model architecture, two-stage policy learning, and real-world deployment method. The policy predicts short-horizon action chunks from egocentric images, UAV state, and natural-language instructions. As shown in Fig.~\ref{fig:method}, the first training stage uses mixed demonstrations from UAV-Flow and UAV-Flow-Sim for imitation learning. The second stage performs online reinforcement learning post-training in simulation suite using Tail-Aware GRPO. It constructs advantages from the group reward distribution to better exploit rare high-reward trajectories, giving trajectories with high completion scores greater influence on policy updates toward higher task completion. 
For real-world deployment on physical hardware, we adopt a ground-drone collaborative architecture. By scheduling execution waypoints according to end-to-end delay, this design bridges discrete policy predictions with continuous closed-loop execution in the real world.

\subsection{Model Architecture}
\label{sec:method_pipeline}

Our policy consists of a pretrained Qwen3-VL~\cite{bai2025qwen3vl} backbone and a lightweight discrete action head. At each time step $t$, the policy receives the initial episode image, the current egocentric image, the UAV state, and a natural-language instruction. The initial image provides a persistent scene reference, while the current image supplies visual feedback for closed-loop control. For simplicity, the UAV state is represented as $(x_t,y_t,z_t,\psi_t)$ in the episode-local trajectory frame, where $\psi_t$ denotes yaw. These state values are serialized as text and concatenated with the language instruction to form the model's text input.

The Qwen3-VL backbone jointly processes visual tokens from the two images, the state-instruction text, and $H$ action tokens appended to the input sequence. The final-layer hidden state of each action token serves as an action query corresponding to one future control step. A lightweight MLP action head maps these queries to action distributions for the future control steps, predicting a chunk of $H$ four-dimensional position-yaw actions in the local coordinate frame, $\mathbf{A}_t=[\mathbf{a}_{t,1},\ldots,\mathbf{a}_{t,H}]$.

To construct the discrete action space, we normalize each action dimension using percentile-based statistics computed from the training set and uniformly divide each normalized dimension into $K$ bins. The lightweight action head outputs categorical logits $\mathbf{Z}_t \in \mathbb{R}^{H \times 4 \times K}$, where $\mathbf{Z}_{t,h,j}$ corresponds to the categorical prediction for the $j$-th action dimension at the $h$-th future step. This discrete representation captures action distributions across UAV maneuvers and provides action log-probabilities for subsequent policy-gradient optimization. Each predicted action chunk forms a short-horizon local motion plan that can be executed by the controller.

\par
\endgroup

\subsection{Hybrid Imitation Pretraining}
\label{sec:method_pretraining}

We pretrain the policy by imitation using a mixture of real-world flight demonstrations from UAV-Flow and simulated demonstrations from UAV-Flow-Sim. Both sources follow a unified episode format and are expanded into frame-level training samples. 
The pretraining dataset comprises approximately 20,000 demonstration trajectories. While real-world demonstrations provide authentic outdoor visual diversity, sensor stochasticity, and physical motion dynamics, simulated demonstrations offer controllable and scalable coverage of Flow maneuvers and task configurations. Jointly training on these complementary data sources enables the policy to map egocentric visual observations, UAV states, and language instructions into future action chunks.

For each training sample, we convert the expert future action chunk into categorical labels using the normalization and discretization, and supervise action prediction with categorical cross-entropy. The loss is averaged over the $H$ future steps and four action dimensions:
\begin{equation}
\mathcal{L}_{\mathrm{CE}} =
\frac{1}{4H}
\sum_{h=1}^{H}\sum_{j=1}^{4}
\mathrm{CE}
\left(
\mathbf{Z}_{t,h,j}, b(a^*_{t,h,j})
\right),
\label{eq:discrete_action_loss}
\end{equation}
where $a^*_{t,h,j}$ is the normalized expert action and $b(\cdot)$ maps it to the corresponding discrete action bin.
The pretrained policy then predicts short-horizon UAV action chunks conditioned on language instructions, serves as the policy for subsequent closed-loop evaluation, and initializes reinforcement learning post-training.

\subsection{RL Post-Training}
\label{sec:method_rl}

Imitation pretraining learns language-conditioned flight control from expert demonstrations, but primarily covers states along expert trajectories. During closed-loop execution, control deviations can shift subsequent observations away from the demonstration distribution, leading to compounding errors. To improve the policy's ability to handle states induced by its own execution, we perform reinforcement learning post-training in UAV-Flow-Sim, further optimizing the policy through environment interaction and task-level feedback.

We use GRPO~\cite{shao2024deepseekmath} as the basis for closed-loop policy optimization. For each task context, the rollout policy $\pi_{\mathrm{old}}$ generates a group of trajectories $\{z_i\}_{i=1}^{G}$, whose relative performance provides the policy update signal. Each trajectory alternates action-chunk prediction and execution until the episode terminates. We evaluate the executed trajectory using the task completion score $C_k$ defined in the benchmark, and denote the task reward for trajectory $i$ by $r_i=C_k(z_i)\in[0,1]$. This score jointly measures how well maneuver execution and the terminal state satisfy the task requirements, allowing group comparisons to distinguish trajectory completion quality.

Standard GRPO constructs trajectory advantages by standardizing rewards within each group:
\begin{equation}
A_i^{\mathrm{GRPO}}=\frac{r_i-\bar r}{s_r},
\label{eq:grpo_advantage}
\end{equation}
where $\bar r$ and $s_r$ are the mean group reward and standard deviation, respectively. However, UAV tasks require coordinated action sequences to execute the instructed maneuver and satisfy the terminal-state requirements. Trajectories sampled in the same task context may achieve different levels of completion, and some groups may contain only a few high-reward trajectories. We therefore seek to learn more effectively from trajectories with higher completion scores, reinforcing the action sequences that better satisfy the task requirements while making fuller use of high-quality outcomes that are sparsely represented within their groups.

To better exploit these rare trajectories, we adopt a reweighting towards high-reward outcomes and construct advantages based on the group reward distribution~\cite{petersen2021deep,walder2025pkpo,tajwar2026maxrl,ramasubramanian2026tailrl}. Specifically, we focus on the upper reward tail and assign weights according to how frequently different reward thresholds are exceeded, giving greater weight to reward intervals reached by fewer trajectories. Incorporating this advantage construction into GRPO yields Tail-Aware GRPO (TGRPO).

Within the current rollout group, following the threshold-based weighting of TailRL~\cite{ramasubramanian2026tailrl}, we define the weight for a trajectory with reward $r_i$ as
\begin{equation}
\omega(r_i)=\int_0^{r_i}
\frac{1}{\sum_{j=1}^{G}\mathbf 1[r_j>u]}\,du,
\label{eq:tgrpo_weight}
\end{equation}
where $u$ is a reward threshold and the denominator counts trajectories whose rewards exceed it. A threshold exceeded by fewer trajectories gives its corresponding reward interval greater weight. Sorting the group rewards as $r_{(1)}\le\cdots\le r_{(G)}$ and setting $r_{(0)}=0$, the integral can be computed exactly as
\begin{equation}
\omega(r_{(i)})=\sum_{\ell=1}^{i}
\frac{r_{(\ell)}-r_{(\ell-1)}}{G-\ell+1}.
\label{eq:tgrpo_sorted_weight}
\end{equation}
We then center the weights within the group to obtain the task advantage:
\begin{equation}
A_i^{\mathrm{task}}=G\left(\omega(r_i)-\frac{1}{G}\sum_{j=1}^{G}\omega(r_j)\right).
\label{eq:tgrpo_task_advantage}
\end{equation}
This construction applies a nonlinear transformation to trajectory rewards based on their group distribution, followed by mean-centering to obtain trajectory advantages.

For an action $\mathbf a$ along a trajectory, let $\mathbf x$ denote the policy input for its originating action chunk together with its position within the chunk. We use $\widehat{\mathrm{KL}}=\log\pi_{\mathrm{old}}(\mathbf a\mid\mathbf x)-\log\pi_{\mathrm{ref}}(\mathbf a\mid\mathbf x)$ to estimate the reference-policy KL penalty, where $\pi_{\mathrm{ref}}$ is the reference policy. These estimates are accumulated along each trajectory and scaled by the KL coefficient. The resulting penalties are centered within the group and subtracted from the task advantages to obtain the final trajectory advantages $A_i$.

Each action probability is the joint probability of its four discretized coordinates. The action probability ratio between the current policy and the rollout policy is
\begin{equation}
\rho=\frac{\pi_\theta(\mathbf a\mid\mathbf x)}
{\pi_{\mathrm{old}}(\mathbf a\mid\mathbf x)}.
\label{eq:tgrpo_ratio}
\end{equation}
We share the final trajectory advantage across all valid actions in that trajectory and retain GRPO's clipped policy objective:
\begin{equation}
\mathcal L_{\mathrm{TGRPO}}
=-\mathbb E\!\left[
\min\!\left(
\rho A,\;
\operatorname{clip}(\rho,1-\epsilon_{\mathrm{low}},1+\epsilon_{\mathrm{high}})A
\right)\right].
\label{eq:grpo_loss}
\end{equation}
Here, $A$ is the final advantage of the trajectory, and $\epsilon_{\mathrm{low}}$ and $\epsilon_{\mathrm{high}}$ specify the clipping bounds. The expectation is estimated with equal weight per trajectory, with each trajectory loss defined as the mean loss over its valid actions.

\Needspace{6\baselineskip}
\subsection{Real-World Deployment}
\label{sec:method_real_world_adaptation}

To deploy the trained policy on a physical UAV, we use a ground-drone collaborative execution architecture, as illustrated in Fig.~\ref{fig:deployment_system}. Due to limited onboard computing resources, VLA inference is performed on the ground station. 
The UAV transmits egocentric images and state estimates to the ground station via a wireless link. Based on the received observations and language instruction, the ground station generates action chunks, converts them into control targets, and transmits them back to the UAV for execution by its onboard flight controller.

The policy outputs position-yaw actions in a local coordinate frame. The execution layer decodes the discrete predictions into continuous actions and incrementally converts each action chunk into position-yaw waypoints using the UAV pose associated with the input observation.

\par
\medskip
\noindent\begin{minipage}[t]{0.47\textwidth}
\vspace{0pt}
Smooth and continuous flight requires the execution layer to account for the latency between observation acquisition and command dispatch. Pausing UAV motion while waiting for inference to complete would introduce interruptions, compromising the continuous execution required by Flow tasks. 
Conversely, allowing the UAV to continue moving during inference may render the initial waypoints in the returned action chunk outdated relative to the UAV's current execution state. To address this issue, we adopt a delay-aware waypoint selection strategy. Let $\delta$ denote the measured delay from image capture to command dispatch, and let $f_c$ denote the execution frequency of the predicted waypoints. The execution layer skips the first $m=\min\left(H-1,\left\lceil\delta f_c\right\rceil\right)$ waypoints in the chunk and dispatches the remaining target.
By compensating for the elapsed time during communication and inference, this mechanism aligns the predicted waypoints with the current execution progress, enabling continuous closed-loop flight.
\end{minipage}\hfill
\begin{minipage}[t]{0.5\textwidth}
\vspace{0pt}
\centering
\includegraphics[width=0.99\linewidth]{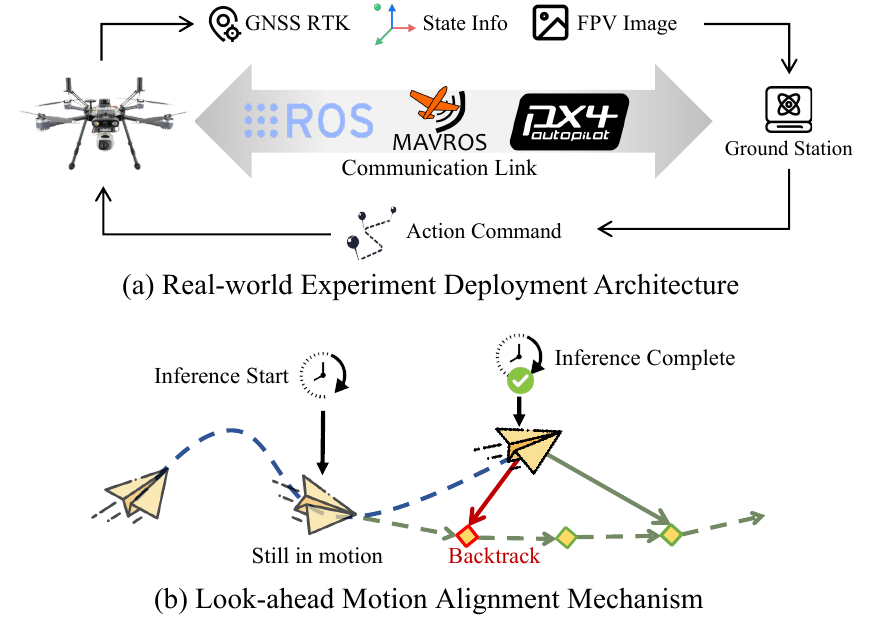}
\captionsetup{skip=12pt,hypcap=false}
    \captionof{figure}{\textbf{Ground-drone collaborative deployment.}
    (a) UAV state estimates and first-person images are transmitted to the ground station, which returns action commands through the communication link. (b) The UAV continues moving during inference; look-ahead waypoint selection accounts for this motion when dispatching the predicted commands.}
    \label{fig:deployment_system}
\end{minipage}
\par

\FloatBarrier
\section{Experiments}
\label{sec:experiments}

We assess FlowVLA through closed-loop simulation, ablation studies, and physical UAV deployment. We first compare the pretrained policy with adapted baselines and evaluate the effects of GRPO and TGRPO post-training on task completion and trajectory similarity. Ablation studies then compare alternative RL algorithms and examine the effects of training-data composition, action representation, and VLM feature selection. Finally, we present flight execution examples in outdoor and indoor real-world scenes.

\subsection{Experimental Setup}
\label{sec:experimental_setup}

\textbf{Training and evaluation dataset.}
The pretraining set comprises 9,766 real-world trajectories selected for task balance and trajectory quality, together with 10,109 simulated trajectories, totaling 19,875 trajectories and 946,607 frames. For closed-loop evaluation, we additionally construct 273 annotated UAV-Flow-Sim test episodes in the same campus scene used to collect the simulated demonstrations. Each episode specifies an initial state, an instruction, a reference trajectory, and task-specific completion criteria.

\textbf{Evaluation protocol and metrics.}
All methods follow the closed-loop evaluation protocol in Sec.~\ref{sec:simulation_suite}, using the same initial state and language instruction for each test episode. Episodes terminate upon task completion, collision, or reaching the maximum horizon; an episode is counted as successful when all applicable task completion criteria are satisfied. We report success rate (SR), completion score (CS), and NDTW, which measure the proportion of successful episodes, the degree of task completion, and the similarity between executed and reference trajectories, respectively.

\begin{table}[!t]
\centering
\uavtablestyle
\caption{\textbf{Closed-loop performance comparison on UAV-Flow-Sim.} We report overall performance and results on the four RL training tasks: Pass, Rotate, Land, and Approach. FlowVLA represents the imitation-pretrained policy; FlowVLA-GRPO and FlowVLA-TGRPO denote its RL post-training variants. Metrics are aggregated over three evaluation runs, reported as mean $\pm$ std for overall performance and as means for individual tasks. Bold text highlights the best Success Rate (SR).}
\label{tab:main_benchmark}

\newcommand{\uavms}[2]{\ensuremath{#1_{\pm #2}}}
\begingroup
\footnotesize
\setlength{\tabcolsep}{1.25pt}
\begin{tabular*}{\textwidth}{@{\extracolsep{\fill}}l*{10}{c}@{}}
\toprule
\multirow{2}{*}[-0.75ex]{\textbf{Method}}
& \multicolumn{2}{c}{\textbf{Overall}}
& \multicolumn{2}{c}{\textbf{Pass}}
& \multicolumn{2}{c}{\textbf{Rotate}}
& \multicolumn{2}{c}{\textbf{Land}}
& \multicolumn{2}{c}{\textbf{Approach}} \\
\cmidrule(lr){2-3}\cmidrule(lr){4-5}\cmidrule(lr){6-7}\cmidrule(lr){8-9}\cmidrule(lr){10-11}
& \textbf{SR (\%)} & \textbf{NDTW}
& \textbf{SR (\%) / CS} & \textbf{NDTW}
& \textbf{SR (\%) / CS} & \textbf{NDTW}
& \textbf{SR (\%) / CS} & \textbf{NDTW}
& \textbf{SR (\%) / CS} & \textbf{NDTW} \\
\midrule
TravelUAV & 56.78 & 0.237 & 42.50 / 0.610 & 0.173 & 46.67 / 0.753 & 0.396 & 59.26 / 0.806 & 0.121 & 47.62 / 0.681 & 0.116 \\
OpenVLA-UAV & 53.85 & 0.345 & 30.00 / 0.513 & 0.239 & 13.33 / 0.527 & 0.269 & 22.22 / 0.493 & 0.118 & 47.62 / 0.581 & 0.243 \\
$\pi_0$-UAV & 68.86 & 0.431 & 57.50 / 0.722 & 0.384 & 46.67 / 0.807 & 0.538 & 68.52 / 0.911 & 0.281 & 54.76 / 0.790 & 0.288 \\
MemoryVLA-UAV & 62.64 & 0.342 & 62.50 / 0.720 & 0.319 & 6.67 / 0.588 & 0.177 & 72.22 / 0.880 & 0.110 & 52.38 / 0.740 & 0.339 \\
$\pi_{0.5}$-UAV & 66.67 & 0.398 & 47.50 / 0.637 & 0.369 & 66.67 / 0.967 & 0.662 & 70.37 / 0.930 & 0.251 & 54.76 / 0.793 & 0.317 \\
\midrule
FlowVLA & \uavms{75.58}{1.12} & \uavms{0.402}{0.006} & 55.00 / 0.716 & 0.394 & 24.44 / 0.540 & 0.334 & 74.69 / 0.878 & 0.305 & 57.94 / 0.813 & 0.355 \\
FlowVLA-GRPO & \uavms{92.06}{0.56} & \uavms{0.434}{0.005} & 85.83 / 0.879 & 0.391 & \textbf{100.00} / 1.000 & 0.660 & 95.06 / 0.994 & 0.301 & 84.13 / 0.952 & 0.335 \\
FlowVLA-TGRPO & \uavms{\mathbf{93.16}}{0.42} & \uavms{0.424}{0.008} & \textbf{90.83} / 0.902 & 0.335 & \textbf{100.00} / 1.000 & 0.660 & \textbf{96.91} / 0.982 & 0.274 & \textbf{90.48} / 0.975 & 0.314 \\
\bottomrule
\end{tabular*}
\endgroup

\end{table}

\textbf{Supervised pretraining.}
Our FlowVLA policy uses a Qwen3-VL backbone and a discrete action head. Conditioned on the initial and current images, UAV state, and instruction, it predicts a ten-step position-yaw action chunk, with 256 bins for each of the four action dimensions. With trajectories sampled at 5Hz, each action chunk spans 2s. Supervised pretraining fine-tunes the VLM through LoRA and trains the action head with categorical cross-entropy on the mixed dataset using the AdamW optimizer. We pretrain FlowVLA for 80,000 steps on eight NVIDIA L40 GPUs. The learning rate is warmed up to $2\times10^{-4}$ over the first 10,000 steps, followed by cosine decay, with gradient clipping set to 1.0.

\textbf{RL post-training.}
Within the SimpleVLA-RL framework~\cite{li2026simplevlarl}, we perform online post-training with GRPO and TGRPO, each initialized from the same pretrained FlowVLA policy. Training covers Pass, Rotate, Land, and Approach, with 204 training trajectories per task and a total of 90 policy update steps. Both methods leverage FSDP across four GPUs to optimize the LoRA parameters and action head with a learning rate of $3\times10^{-5}$, using a PPO clipping range of $\epsilon_{\mathrm{low}}=\epsilon_{\mathrm{high}}=0.2$. At each RL iteration, we sample 8 closed-loop rollout trajectories per task across 16 parallel task configurations at temperature $T=1.6$, yielding 128 rollout trajectories per iteration with a maximum horizon of 100 simulator interactions per rollout. We then perform one optimization epoch over the collected batch with a global mini-batch size of 16 trajectories, executing 8 optimizer steps per iteration. At the end of each episode, the trajectory-level completion score serves as terminal feedback for advantage construction and policy updates. Greedy decoding is employed during evaluation.

\begin{figure}[!t]
\centering
\includegraphics[width=\textwidth]{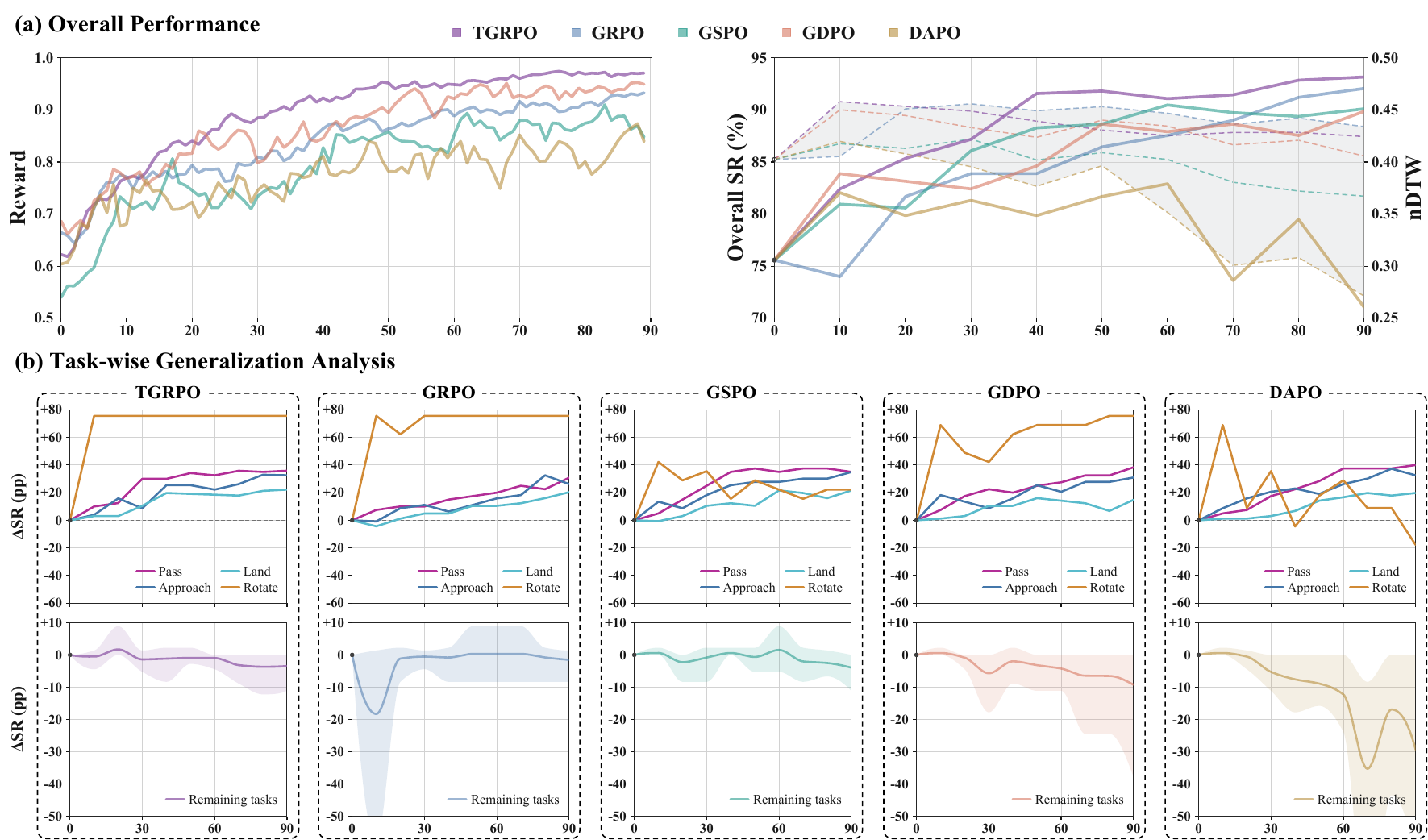}
\caption{\textbf{Performance comparison of RL post-training algorithms on UAV-Flow-Sim.} 
(a) Left: mean within-group training reward against policy update steps. Right: overall test-set metrics over time, showing SR as solid lines on the left axis and nDTW as dashed lines on the right axis.
(b) SR gains relative to the pretrained baseline in percentage points. The top row reports performance across the four RL training tasks, whereas the bottom row displays the mean gain over the remaining tasks excluded from post-training, with shaded regions denoting the min--max range across these tasks.}
\label{fig:rl_algorithm_ablation}
\vspace{-4pt}
\end{figure}

\subsection{Baseline Models}
\label{sec:baseline_models}

We select five representative methods as baselines and adapt them to the Flow task. All baselines are trained by imitation learning on the same simulated demonstrations, with their VLM components fine-tuned using LoRA. To provide a common closed-loop execution interface, each model's outputs are converted to local position-yaw commands and executed by the same simulator-side controller.

TravelUAV~\cite{wang2024openuav} uses a hierarchical architecture: a multimodal language model predicts a target pose from the current observation, UAV state, and task instruction, followed by a fine-grained trajectory decoder that generates a local flight trajectory. OpenVLA-UAV~\cite{kim2024openvla} autoregressively predicts discretized action tokens from visual observations and language instructions. MemoryVLA-UAV~\cite{shi2025memoryvla} maintains separate memory banks for historical visual features and instruction-conditioned semantic features, combining retrieved memories with current observation features to condition a diffusion model for action-sequence generation. $\pi_0$-UAV~\cite{black2024pi0} generates continuous action chunks through a flow-matching action expert conditioned on VLM features. $\pi_{0.5}$-UAV~\cite{physicalintelligence2025pi05} builds on this framework with heterogeneous robot data and semantic supervision to improve cross-task generalization.

FlowVLA denotes the policy obtained by imitation pretraining on mixed real-world and simulated demonstrations. FlowVLA-GRPO and FlowVLA-TGRPO denote variants that undergo RL post-training from this policy using the corresponding algorithm. 
To examine both optimization efficiency and performance on these tasks, we evaluate all policies on the full test set, quantifying performance improvements on the four RL training tasks alongside capability retention on the remaining tasks.

\subsection{Closed-Loop Simulation Results}
\label{sec:closed_loop_results}

\textbf{Overall performance.}
Table~\ref{tab:main_benchmark} shows that our pretrained policy achieves a mean overall SR of 75.58\%, exceeding the strongest evaluated baseline $\pi_0$-UAV at 68.86\%, by 6.72 percentage points. Its mean overall NDTW of 0.402 is lower than $\pi_0$-UAV's 0.431 but higher than those of the remaining baselines. These results demonstrate that FlowVLA achieves a superior overall success rate compared to existing baselines, providing a strong initialization for subsequent reinforcement learning post-training.

The baselines exhibit different strengths across maneuver tasks. MemoryVLA-UAV achieves the highest baseline SR on Pass and Land, while $\pi_{0.5}$-UAV performs best on Rotate. Pretrained FlowVLA outperforms all baselines on Land and Approach but still trails the best baselines on Pass and Rotate, leaving substantial room for improvement on these tasks. To further improve task completion, we perform online RL post-training from this pretrained policy and compare the effects of GRPO and TGRPO.

\textbf{RL post-training performance.}
As shown in Table~\ref{tab:main_benchmark}, FlowVLA-GRPO and FlowVLA-TGRPO, initialized from the same pretrained policy, both improve overall task completion, achieving mean overall SRs of 92.06\% and 93.16\%, respectively. FlowVLA-TGRPO raised the gain to 17.58 percentage points over the pretrained policy and by a \mbox{further} 1.10  percentage points over FlowVLA-GRPO. These improvements extend across all four RL training tasks: FlowVLA-TGRPO increases SR by more than 20 percentage points each of Pass, Rotate, Land, and Approach, achieving over 90\% SR on all four tasks. 
These RL-refined models achieve mean overall nDTW scores of 0.434 and 0.424, respectively, both outperforming the pretrained policy. These results suggest that RL post-training enhances end-task completion while simultaneously improving alignment with the reference trajectories.
Task-wise results further show that FlowVLA-TGRPO's gains over FlowVLA-GRPO are more pronounced on Pass and Approach.  With the same policy initialization and training configuration, these results support the effectiveness of the tail-aware advantage in further improving task success rates.

\begin{table}[!t]
\centering
\uavtablestyle
\caption{\textbf{Comparison of RL algorithms on UAV-Flow-Sim.} All variants start from the same pretrained FlowVLA policy. We report overall performance and results on the four RL training tasks: Pass, Rotate, Land, and Approach. Overall metrics are reported as mean $\pm$ standard deviation across three selected evaluations per variant; task-wise metrics are means. Bold denotes the highest SR.}
\label{tab:rl_optimizer_ablation}
\begingroup
\footnotesize
\setlength{\tabcolsep}{1.25pt}
\begin{tabular*}{\textwidth}{@{\extracolsep{\fill}}l*{10}{c}@{}}
\toprule
\multirow{2}{*}[-0.75ex]{\textbf{RL Algorithm}}
& \multicolumn{2}{c}{\textbf{Overall}}
& \multicolumn{2}{c}{\textbf{Pass}}
& \multicolumn{2}{c}{\textbf{Rotate}}
& \multicolumn{2}{c}{\textbf{Land}}
& \multicolumn{2}{c}{\textbf{Approach}} \\
\cmidrule(lr){2-3}\cmidrule(lr){4-5}\cmidrule(lr){6-7}\cmidrule(lr){8-9}\cmidrule(lr){10-11}
& \textbf{SR (\%)} & \textbf{NDTW}
& \textbf{SR (\%) / CS} & \textbf{NDTW}
& \textbf{SR (\%) / CS} & \textbf{NDTW}
& \textbf{SR (\%) / CS} & \textbf{NDTW}
& \textbf{SR (\%) / CS} & \textbf{NDTW} \\
\midrule
DAPO & $82.91_{\pm1.18}$ & $0.352_{\pm0.031}$ & 92.50 / 0.912 & 0.354 & 53.33 / 0.700 & 0.181 & 91.36 / 0.954 & 0.245 & 84.13 / 0.917 & 0.306 \\
GSPO & $89.13_{\pm1.29}$ & $0.394_{\pm0.007}$ & 91.67 / 0.907 & 0.299 & 40.00 / 0.698 & 0.527 & \textbf{95.68} / 0.968 & 0.273 & 85.71 / 0.965 & 0.263 \\
GDPO & $89.87_{\pm2.44}$ & $0.406_{\pm0.018}$ & \textbf{93.33} / 0.926 & 0.321 & \textbf{100.00} / 1.000 & 0.664 & 89.51 / 0.876 & 0.230 & \textbf{88.89} / 0.944 & 0.300 \\
GRPO & $\mathbf{92.06}_{\pm0.56}$ & $0.434_{\pm0.005}$ & 85.83 / 0.879 & 0.391 & \textbf{100.00} / 1.000 & 0.660 & 95.06 / 0.994 & 0.301 & 84.13 / 0.952 & 0.335 \\
\bottomrule
\end{tabular*}
\endgroup
\end{table}

\begin{table}[!t]
\centering
\uavtablestyle
\caption{\textbf{Ablation of training data and model variants on UAV-Flow-Sim.} Results are reported as mean $\pm$ standard deviation across three evaluations.}
\label{tab:head_ablation}
\begin{tabular*}{0.6\columnwidth}{@{\extracolsep{\fill}}lccc@{}}
\toprule
\textbf{Variant} & \textbf{Data} & \textbf{SR (\%)} & \textbf{NDTW} \\
\midrule
FlowVLA & Sim & $73.14_{\pm0.56}$ & $0.399_{\pm0.016}$ \\
FlowVLA & Mixed & $75.58_{\pm1.12}$ & $0.402_{\pm0.006}$ \\
FlowVLA w/ FAST head & Mixed & $64.10_{\pm0.63}$ & $0.340_{\pm0.007}$ \\
FlowVLA w/ all layers & Mixed & $76.19_{\pm0.73}$ & $0.419_{\pm0.022}$ \\
\bottomrule
\end{tabular*}

\end{table}

\textbf{RL training dynamics.}
Fig.~\ref{fig:rl_algorithm_ablation}(a) further illustrates the training processes of GRPO and TGRPO. Compared with GRPO, TGRPO's mean within-group reward rises faster during early training and gradually increases to approximately 0.97 with subsequent updates. This trend is also reflected in task success on the test set: FlowVLA-TGRPO achieves an overall SR above 90\% after approximately 40 updates, whereas FlowVLA-GRPO reaches this level after approximately 80 updates, indicating that FlowVLA-TGRPO requires fewer policy updates to achieve a high task success rate. Meanwhile, the nDTW metric exhibits early gains for both methods before entering a phase of slight fluctuation as post-training progresses.

\subsection{Ablation Studies}
\label{sec:ablation_studies}

\textbf{Comparison of RL algorithms.}
To examine the effects of different RL algorithms on post-training, we further compare GRPO with DAPO~\cite{yu2025dapo}, GSPO~\cite{zheng2025gspo}, and GDPO~\cite{liu2026gdpo}, all initialized from the same pretrained FlowVLA policy. GDPO and GRPO use the same clipping parameters and both apply KL regularization relative to a reference policy. DAPO and GSPO instead set the upper clipping bound to 0.28 and do not use reference-policy KL regularization. As shown in Table~\ref{tab:rl_optimizer_ablation}, GRPO achieves an overall SR of 92.06\% and an overall NDTW of 0.434, both the highest among the four methods.

The training curves in Fig.~\ref{fig:rl_algorithm_ablation}(a) further illustrate how these methods perform over the course of training. Overall SR generally increases for FlowVLA-GRPO and FlowVLA-GDPO, while GSPO remains near 90\% in later training and DAPO declines substantially. FlowVLA-GRPO maintains relatively high NDTW in later training, whereas FlowVLA-GDPO and FlowVLA-GSPO show declines. The late-stage performance degradation of FlowVLA-DAPO and FlowVLA-GSPO may be partly related to their lack of reference-policy KL regularization, as their updates do not explicitly constrain cumulative deviation from the pretrained reference policy.

Fig.~\ref{fig:rl_algorithm_ablation}(b) further distinguishes performance changes on the four RL training tasks from those on the remaining tasks. FlowVLA-GRPO substantially improves all four training tasks while recovering from an early decline on the remaining tasks and returning close to the pretrained level by the end of training. FlowVLA-TGRPO also improves all four training tasks and reaches a high success rate on Rotate early in training, but its mean SR on the remaining tasks declines slightly during later training. GSPO likewise shows a small decline on the remaining tasks, whereas GDPO and DAPO exhibit larger declines. Together with the overall results above, these findings show that FlowVLA-GRPO offers a favorable combination of task completion, reference-trajectory similarity, and retention of pretrained capabilities, supporting its use as the basis for further improvements to advantage computation. FlowVLA-TGRPO achieves a higher overall success rate. Although its performance loss on the remaining tasks is slightly greater than GRPO's, it is substantially smaller than those of GDPO and DAPO, indicating that TGRPO further improves completion of the training tasks while largely retaining pretrained capabilities.

\textbf{Effect of mixed training data.}
As shown in Table~\ref{tab:head_ablation}, adding real-world flight demonstrations increases FlowVLA's mean SR from 73.14\% to 75.58\%, a gain of 2.44 percentage points, while mean NDTW also increases slightly from 0.399 to 0.402. 
These findings indicate that beyond facilitating direct real-world policy learning, physical flight demonstrations provide complementary training signals to simulated data, thereby enhancing closed-loop task success in simulation.

\textbf{Model variants.}
We further replace the discrete action head with a FAST head trained on the same mixed data. FAST compresses quantized frequency-domain representations of action chunks into token sequences for autoregressive prediction~\cite{pertsch2025fast}. With this replacement, mean SR and NDTW are 64.10\% and 0.340, respectively, both below the default head's 75.58\% and 0.402.
We also examine the effect of fusing features from multiple VLM layers on action prediction, using an ELMo-style scalar mix~\cite{peters2018elmo} to form a weighted combination of all Qwen3-VL hidden states, including the embedding output. This variant achieves a mean SR of 76.19\% and a mean NDTW of 0.419, improving over the default final-layer features by 0.61 percentage points and 0.017, respectively. Given the small gain in task success rate, we retain the model using only final-layer features as the common initialization for all RL post-training experiments.

\Needspace{10\baselineskip}
\subsection{Real-World Deployment}
\label{sec:real_world_deployment_exp}

Following the deployment scheme in Sec.~\ref{sec:method_real_world_adaptation}, we deploy VLA policy on a physical UAV and evaluate its ability to execute language instructions in closed-loop flight in outdoor and indoor scenes. During flight, the UAV continuously transmits egocentric images and state estimates to the ground station, which predicts action chunks. The execution layer converts the predictions into waypoint commands, selects look-ahead waypoints based on the end-to-end delay, and dispatches them to the flight controller, maintaining continuous closed-loop flight despite communication and policy-inference delays.

\begin{figure}[!htbp]
\centering
    \includegraphics[width=0.90\textwidth]{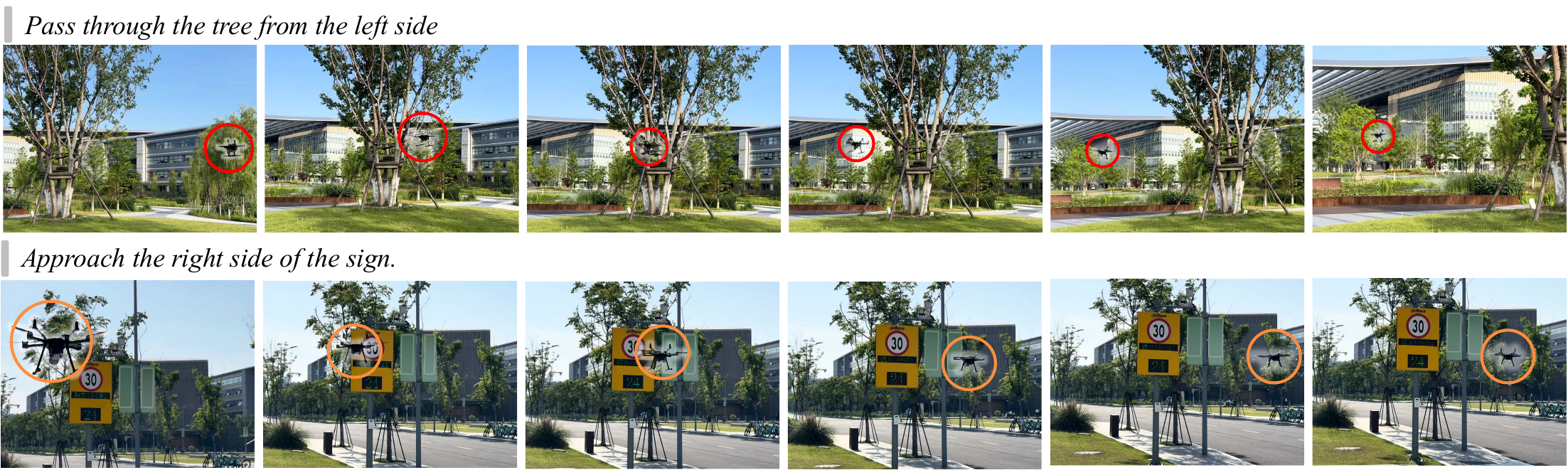}
    \caption{\textbf{Outdoor flight demonstrations with \(\pi_0\)-UAV.}
    Third-person image sequences show passing a tree from the left side and approaching the right side of a sign. Colored circles indicate the UAV position in each frame.}
    \label{fig:deploydemo}

\vspace{6pt}
\includegraphics[width=0.90\textwidth]{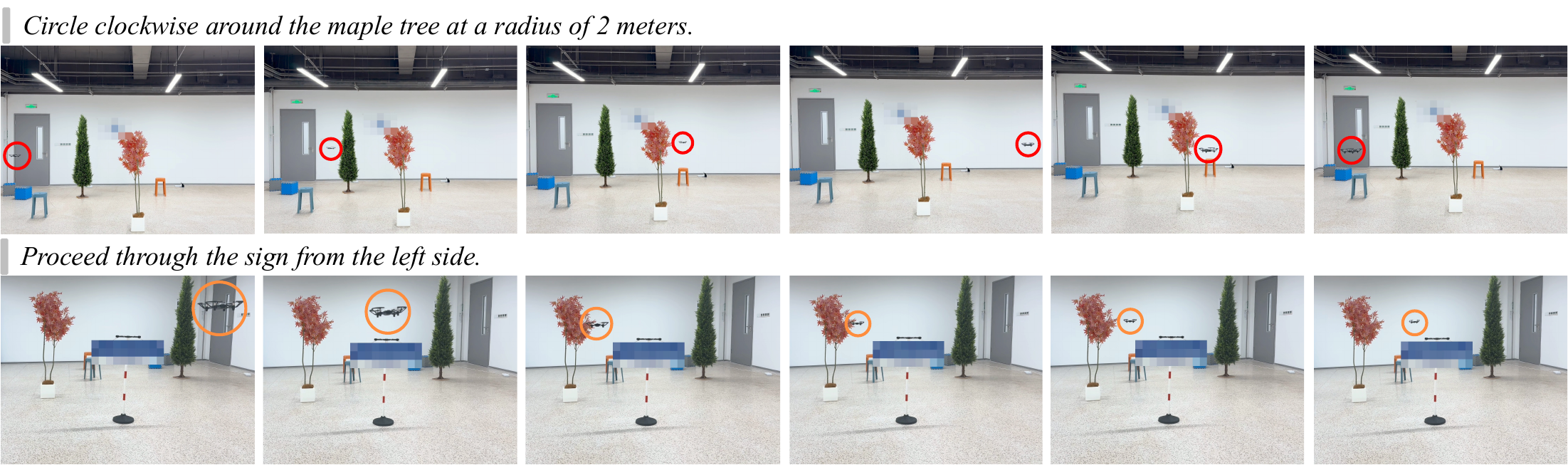}
    \caption{\textbf{Indoor UAV deployment examples with FlowVLA policy.}
    Third-person image sequences show clockwise orbiting around a tree and passing a sign from the left side. Colored circles indicate the UAV position in each frame.}
    \label{fig:indoor_demo}
\end{figure}

Fig.~\ref{fig:deploydemo} presents outdoor executions by $\pi_0$-UAV, demonstrating passing a tree from the left side and approaching the right side of a sign. Meanwhile, Fig.~\ref{fig:indoor_demo} illustrates indoor deployments of FlowVLA, featuring clockwise orbiting around a tree and passing a sign from the left side. 
Covering fundamental passing, approaching, and orbiting maneuvers, these indoor and outdoor sequences validate the end-to-end policies' ability to localize target objects from visual observations and map spatial instruction constraints into continuous flight actions. 
Collectively, these physical deployments highlight the policies' robust object-relative navigation capabilities, substantiating the quantitative evaluation from our closed-loop simulations.

\FloatBarrier
\Needspace{6\baselineskip}

\section{Conclusion}

We introduce Flying-on-a-Word (Flow) to study UAV execution of short-range maneuvers from natural-language instructions, egocentric observations, and the current state. To support this task, we construct UAV-Flow, a real-world dataset of 30,692 expert-piloted trajectories, and UAV-Flow-Sim, which provides 10,109 simulated demonstrations and a closed-loop evaluation environment. Building on these resources, FlowVLA undergoes imitation pretraining on mixed real-world and simulated demonstrations, followed by online post-training in simulation with Tail-Aware GRPO (TGRPO). 
Across 273 held-out UAV-Flow-Sim episodes, FlowVLA pretrained on mixed demonstrations achieves a mean success rate of 75.58\%. Subsequent TGRPO post-training further improves the success rate to 93.16\% with substantially fewer training updates, demonstrating faster convergence and improved sample efficiency. 
Despite these strong simulation results, key practical limitations remain. Current high-rate closed-loop control relies heavily on ground-station computation and wireless communication, introducing latency risks and constraining real-time onboard autonomy. In future work, we will focus on model compression and edge-efficient deployment to enable low-latency onboard inference, such as quantization, distillation, and optimized architecture design. We will also scale up and diversify the training data across environments and tasks to improve generalization to unseen scenes and instructions. Furthermore, we will explore policy optimization strategies that jointly balance task completion, trajectory quality, and the preservation of previously learned capabilities.

\Needspace{8\baselineskip}
\section*{Acknowledgments}

This research is supported in part by the National Key R\&D Program of China (Grant No. 2024YFF1401400), the Science Fund for Creative Research Groups (No. T2621006), the National Natural Science Foundation of China (Nos. U23B2010, 62461160308, 62576024), the Fundamental Research Funds for the Central Universities (No. 501RCQD2025141003), and BeiHang GanWei Project (No. 502GWXM2024141001).

\FloatBarrier
{\small
\setstretch{1.0}
\fontsize{8.5}{10.2}\selectfont
\setlength{\bibsep}{1.5pt plus 0.3pt}
\interlinepenalty=10000
\bibliographystyle{\colabbibstyle}
\bibliography{ref}
}

\clearpage
\appendix
\pdfbookmark[0]{Appendix}{appendix}
\section*{Appendix}
\section{Reward Components for RL Tasks}
\label{supp:evaluation}

The task-level completion score $C_k$ supplies terminal feedback for RL post-training. Its components are defined separately for the four tasks.

\textbf{Pass.}
The reward has three components: a passage reward of 0.5 for entering the corridor on the requested side, a terminal-position reward of up to 0.5 based on distance to the reward region, and a collision penalty of $-0.3$.

\textbf{Rotate.}
The reward has two components: a rotation-accuracy reward of up to 0.7 for matching the required cumulative yaw change, and a position-retention reward of up to 0.3 that penalizes drift from the initial position.

\textbf{Land.}
The reward has three components: an approach-region reward of up to 0.5 based on the terminal distance to the requested landing region, a descent reward of 0.5 when the vertical descent exceeds the specified threshold, and a collision penalty of $-0.3$.

\textbf{Approach.}
The reward has two components: a terminal-region reward of up to 1.0 based on the distance to the requested object-relative region, and a collision penalty of $-0.3$.

Using the interval score $\Phi(d;\Delta,w)$ defined in the main paper, the four RL tasks have the following completion scores. Distances are in centimeters and angles are in degrees.
\begin{equation}
\begin{aligned}
C_{\rm Pass}&=\mathcal C\bigl(0.5b+\Phi(d_P;200,0.5)-0.3c\bigr),\\
C_{\rm Rotate}&=\mathcal C\bigl(\Omega(e_R)+\Phi(d_R;100,0.3)\bigr),\\
C_{\rm Land}&=\mathcal C\bigl(\Phi(d_L;50,0.5)
 +0.5\,\mathbf1[\delta z>80]-0.3c\bigr),\\
C_{\rm Approach}&=\mathcal C\bigl(\Phi(d_A;50,1)-0.3c\bigr),
\end{aligned}
\label{supp:eq:scores}
\end{equation}
where $\mathcal C(u)=\operatorname{clip}(u,0,1)$, $b$ indicates whether the UAV has entered the Pass corridor, $c$ indicates a collision with the target footprint, and $\delta z$ is the descent in world coordinates. The distances $d_P$, $d_L$, and $d_A$ are measured from the terminal XY position to the corresponding reward regions, with zero distance inside or on the boundary. These reward regions are defined by the target object's position and the instructed spatial relation. For Rotate, $d_R$ measures initial-to-final positional drift beyond the tolerance, $e_R$ measures the deviation of the cumulative yaw change from its target, and $\Omega(e_R)$ is the corresponding rotation-accuracy score.
\section{Additional Training Settings}
\label{supp:policy}

\textbf{Imitation pretraining.}
Table~\ref{supp:tab:sft} supplements the pretraining settings in the main paper. The LoRA adapters and action head share the same optimizer and learning rate.

\begin{table}[!htbp]
\centering\supptablestyle
\caption{\textbf{Additional imitation-pretraining settings.}}
\label{supp:tab:sft}
\begin{tabularx}{0.68\linewidth}{Y l}
\toprule
Parameter & Value \\
\midrule
Backbone & Qwen3-VL-2B-Instruct \\
Optimizer & AdamW \\
Adam betas & $(0.9,0.95)$ \\
Weight decay & $10^{-8}$ \\
Minimum learning rate & $5\times10^{-7}$ \\
Global batch size & $128$ \\
LoRA rank / alpha / dropout & $32$ / $16$ / $0.05$ \\
Gradient norm limit & $1.0$ \\
\bottomrule
\end{tabularx}
\end{table}

\textbf{RL post-training.}
Each method starts from the same pretrained policy and uses 204 training trajectories for each of Pass, Rotate, Land, and Approach. Table~\ref{supp:tab:rlconfig} summarizes the optimization and rollout settings. The LoRA adapters and action head are updated with a learning rate of $3\times10^{-5}$. Rollout collection uses categorical sampling to obtain eight trajectories per context, while evaluation selects the most probable action bins. The reported CS retains its original scale. GRPO, TGRPO, and GDPO additionally use reference-policy KL regularization. The sequence ratio, reward-component signal, and reference-policy penalty are described below.

\begin{table}[!htbp]
\centering\supptablestyle
\caption{\textbf{RL optimization and rollout hyperparameters.}}
\label{supp:tab:rlconfig}
\begin{tabularx}{0.68\linewidth}{Y l}
\toprule
Parameter & Value \\
\midrule
Learning rate & $3\times10^{-5}$ \\
Policy mini-batch size & $16$ \\
Rollouts per context & $8$ \\
Sampling temperature & $1.6$ \\
Maximum interactions per rollout & $100$ \\
Lower clipping width & $0.2$ \\
Upper clipping width (GRPO, TGRPO, GDPO) & $0.2$ \\
Upper clipping width (DAPO, GSPO) & $0.28$ \\
KL coefficient (GRPO, TGRPO, GDPO) & $0.001$ \\
Entropy coefficient & $0$ \\
LoRA rank / alpha & $32$ / $16$ \\
GPU count & $4$ \\
\bottomrule
\end{tabularx}
\end{table}

\FloatBarrier
\label{supp:rl}

\textbf{GSPO.}
GSPO formulates the clipped policy objective using a shared importance ratio for each action sequence. For a sequence containing $N$ valid action positions, the length-normalized ratio is
\begin{equation}
s=\left(\prod_{n=1}^{N}\rho_n\right)^{1/N}
=\exp\!\left(\frac{1}{N}\sum_{n=1}^{N}\log\rho_n\right),
\label{supp:eq:gspo}
\end{equation}
where $\rho_n$ denotes the action-position probability ratio defined in the main paper. Clipping this shared ratio makes the update depend on the aggregate probability change across the sequence.

\textbf{GDPO.}
Let $r_i^{(j)}$ be reward component $j$ of rollout $i$. The component-normalized signal, using an unweighted sum, is
\begin{equation}
z_i^{(j)}=\frac{r_i^{(j)}-\mu_j}{\sigma_j+\varepsilon},
\qquad A_i^{\mathrm{comp}}=\sum_j z_i^{(j)}.
\label{supp:eq:gdpo}
\end{equation}
Statistics are computed within each task-context group, with $\varepsilon>0$ for stability. Reward components follow the task definitions in Appendix~\ref{supp:evaluation}. $A_i^{\mathrm{comp}}$ denotes the reward signal before any subsequent advantage processing.

\textbf{KL regularization.}
GRPO, TGRPO, and GDPO use KL regularization relative to a reference policy. For GRPO, the regularized trajectory return is
\begin{equation}
\begin{aligned}
\widetilde R_i&=R_i-\beta K_i,\\
K_i&=\sum_{n=1}^{N_i}
\log\frac{\pi_{\mathrm{old}}(\mathbf a_{i,n}\mid\mathbf x_{i,n})}
{\pi_{\mathrm{ref}}(\mathbf a_{i,n}\mid\mathbf x_{i,n})},
\end{aligned}
\label{supp:eq:kl}
\end{equation}
where $\pi_{\mathrm{ref}}$ denotes the reference policy and $\beta$ controls the regularization strength.
\section{Additional Visualizations}
\label{supp:analysis}

\subsection{Trajectory examples}
\begin{figure}[!htbp]
\centering
\includegraphics[width=0.90\textwidth]{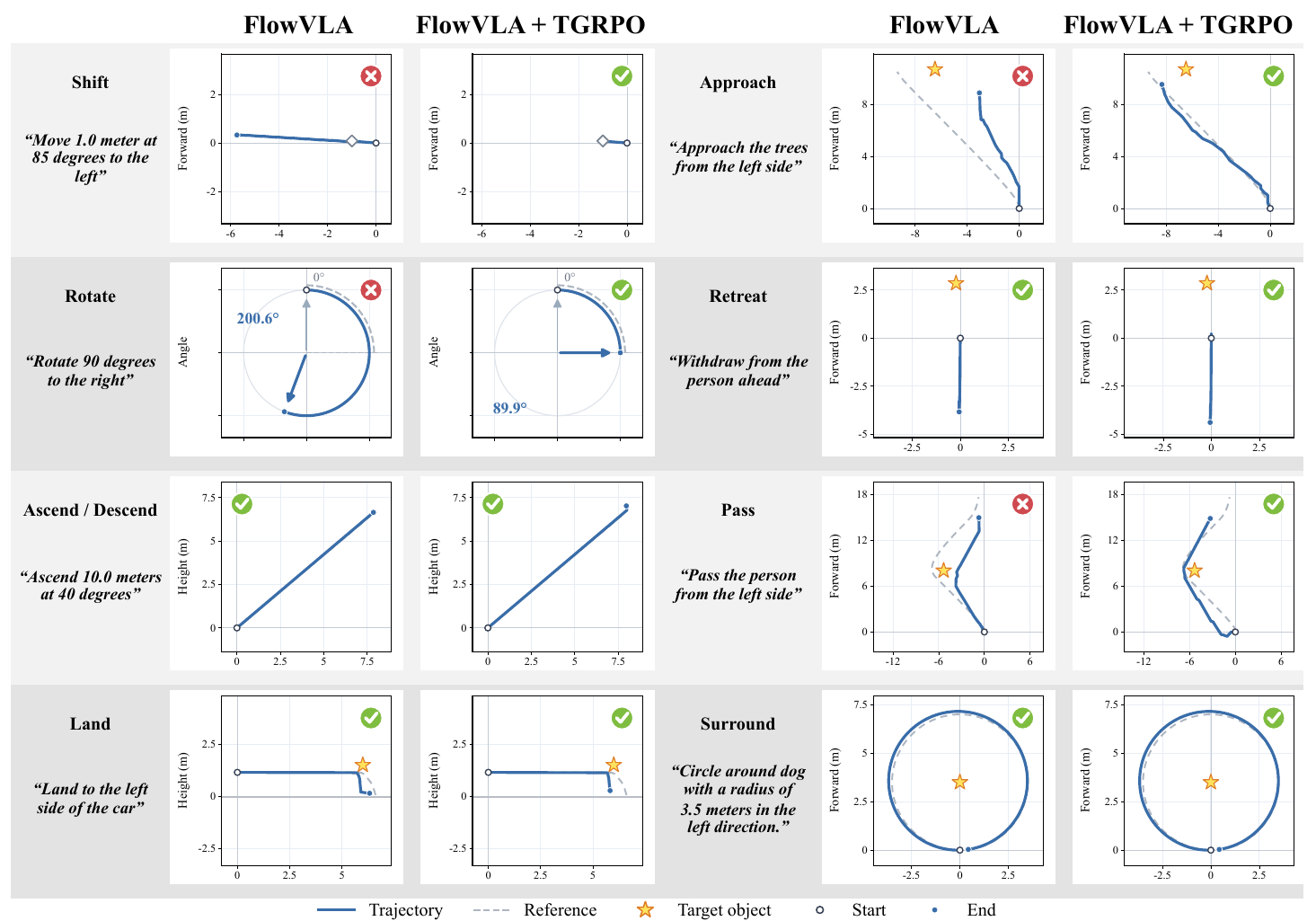}
\caption{\textbf{Paired closed-loop trajectories} for Shift, Rotate, Ascend/Descend, Land, Approach, Retreat, Pass, and Surround. Each task includes an instruction and adjacent plots for FlowVLA and FlowVLA + TGRPO. Solid blue lines denote executed trajectories, and dashed lines denote references. Stars mark target objects, open circles mark initial states, and filled blue dots mark endpoints. Green check marks and red crosses indicate successful and failed episodes, respectively. Spatial coordinates are in meters. The Rotate plots show clockwise yaw change relative to the initial heading, with arrows indicating the final headings and annotations giving the executed rotation angles.}
\label{supp:fig:legacy}

\end{figure}

Figure~\ref{supp:fig:legacy} compares FlowVLA and its TGRPO-post-trained policy on paired test episodes covering eight task types. In Shift, TGRPO more closely follows the requested displacement and succeeds where FlowVLA fails. In Rotate, TGRPO executes a clockwise rotation of $89.9^\circ$ for a $90^\circ$ instruction, correcting FlowVLA's overshoot to $200.6^\circ$. In Pass, TGRPO successfully passes the target on the requested side, whereas FlowVLA fails. In Approach, TGRPO also succeeds where FlowVLA fails. Both policies succeed in the illustrated Retreat, Ascend/Descend, Land, and Surround episodes. These examples illustrate how TGRPO can improve motion accuracy and adherence to spatial instructions.

\subsection{Physical UAV platforms}
\label{supp:deployment}

\begin{figure}[!htbp]
\centering
\includegraphics[width=0.62\textwidth]{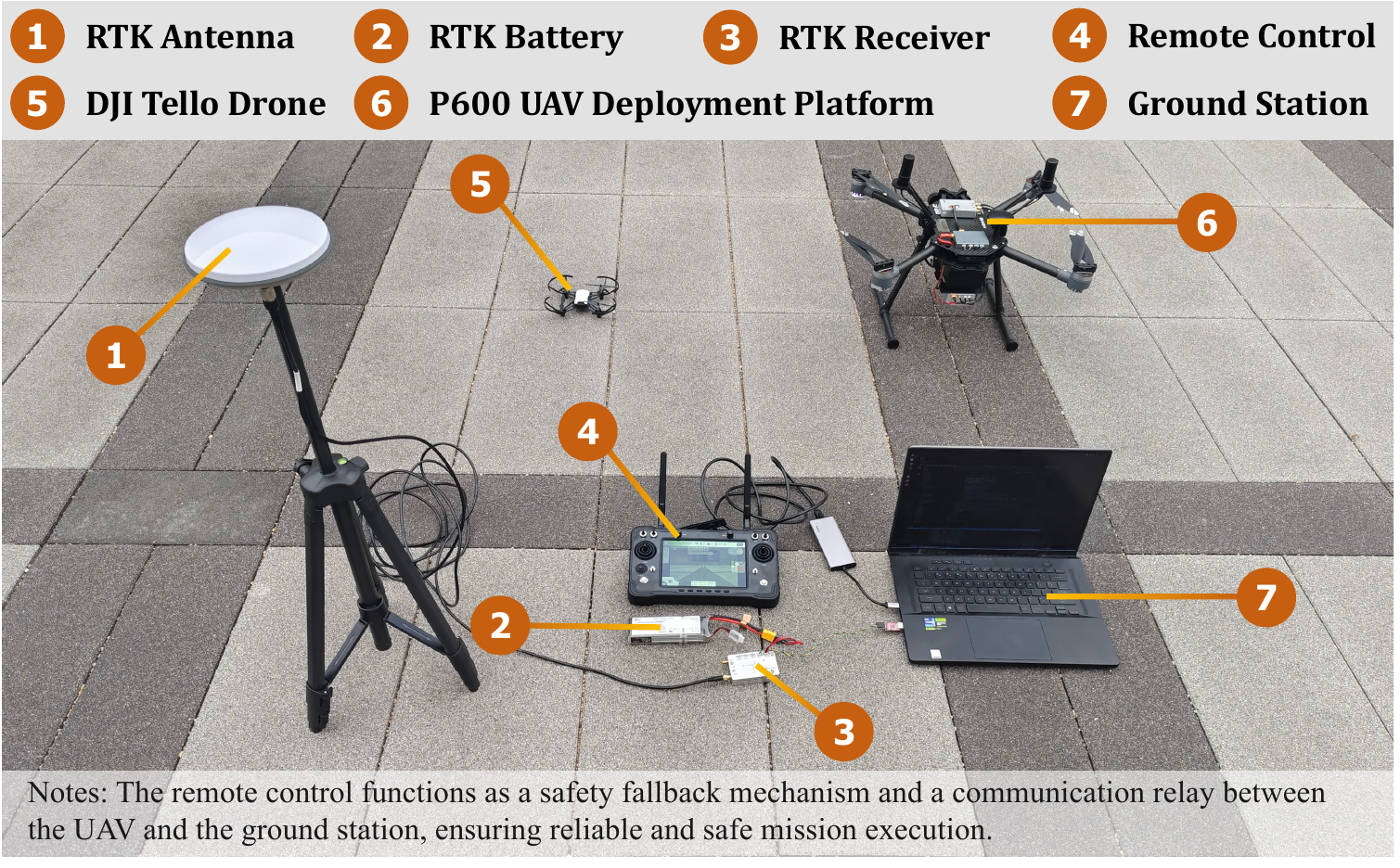}
\caption{\textbf{Hardware setup for physical-flight tests}, including the P600 UAV platform, DJI Tello, RTK equipment, remote control, and ground station~\cite{wang2026uav}.}
\label{supp:fig:platform}
\end{figure}

We test physical execution in indoor and outdoor environments using two UAV platforms: a P600 UAV platform and a DJI Tello, as shown in Fig.~\ref{supp:fig:platform}.
\FloatBarrier

\end{document}